%% file: main.tex
\documentclass[runningheads]{llncs}

\usepackage{eccv}

\usepackage{eccvabbrv}

\usepackage{graphicx}
\usepackage{booktabs}
\usepackage{bm}

\usepackage{tabularx}
\usepackage{algorithm}
\usepackage{algpseudocode}

\usepackage[accsupp]{axessibility}  

\usepackage{hyperref}

\usepackage{orcidlink}

\begin{document}

\newcommand{\rcvehicles}{RC-Vehicles}
\newcommand{\ours}{InstructioNet}

\title{Learning to Build by \\ Building Your Own Instructions} 


\author{
Aaron Walsman\inst{1} \and
Muru Zhang\inst{1} \and
Adam Fishman\inst{1} \and
Ali Farhadi\inst{1} \and
Dieter~Fox\inst{1,2}
}

\authorrunning{A.~Walsman et al.}

\institute{University of Washington
\email{awalsman@cs.washington.edu}\\ \and
NVIDIA
}

\maketitle

\input{sections/abstract}

\input{sections/introduction}

\input{sections/related_work}

\input{sections/method}

\input{sections/experiments}

\input{sections/conclusion}


%
%
\bibliographystyle{splncs04}
\bibliography{main}

\appendix

\input{appendices/additional_ablations}
\input{appendices/expert_details}
\input{appendices/training_algorithm}

\end{document}

%% file: sections/abstract.tex
\begin{abstract}

Structural understanding of complex visual objects is an important unsolved component of artificial intelligence.
To study this, we develop a new technique for the recently proposed Break-and-Make problem in LTRON where an agent must learn to build a previously unseen LEGO assembly using a single interactive session to gather information about its components and their structure.  We attack this problem by building an agent that we call \textbf{\ours} that is able to make its own visual instruction book.
By disassembling an unseen assembly and periodically saving images of it, the agent is able to create a set of instructions so that it has the information necessary to rebuild it.
These instructions form an explicit memory that allows the model to reason about the assembly process one step at a time, avoiding the need for long-term implicit memory.  This in turn allows us to train on much larger LEGO assemblies than has been possible in the past.
To demonstrate the power of this model, we release a new dataset of procedurally built LEGO vehicles that contain an average of 31 bricks each and require over one hundred steps to disassemble and reassemble.  We train these models using online imitation learning which allows the model to learn from its own mistakes.  Finally, we also provide some small improvements to LTRON and the Break-and-Make problem that simplify the learning environment and improve usability.  This data and updated environments can be found at \href{https://github.com/aaronwalsman/ltron/blob/v1.1.0}{github.com/aaronwalsman/ltron/blob/v1.1.0}.  Additional training code can be found at \href{https://github.com/aaronwalsman/ltron-torch/tree/eccv-24}{github.com/aaronwalsman/ltron-torch/tree/eccv-24}.
\end{abstract}

%% file: sections/introduction.tex
\section{Introduction}
The ability to understand and execute complex assembly problems is one of the hallmarks of human intelligence.  Humans use this ability to construct tools and reverse engineer previously unseen part-based objects.

The recently proposed Break-and-Make problem\cite{ltron} is designed to train agents to develop these abilities using complex LEGO structures.  In this problem, an agent must learn to build a previously unseen assembly by actively inspecting it.  To do this the agent is given access to an interactive simulator that allows it to disassemble the structure in order to reveal hidden components and see how everything fits together.  Once it is confident that it knows the structure, the agent is presented with an empty scene and must build the model again from scratch.
This problem is designed to simulate a reverse engineering problem.  The agent must take apart a complex structure in order to learn how to make it.  By training agents to effectively reverse engineer these systems, we can discover new tools for understanding and building intelligent systems that can reason about complex structures.

\begin{figure*}[t!]
  \centering
  \makebox[\columnwidth][c]{\includegraphics[width=1.0\textwidth]{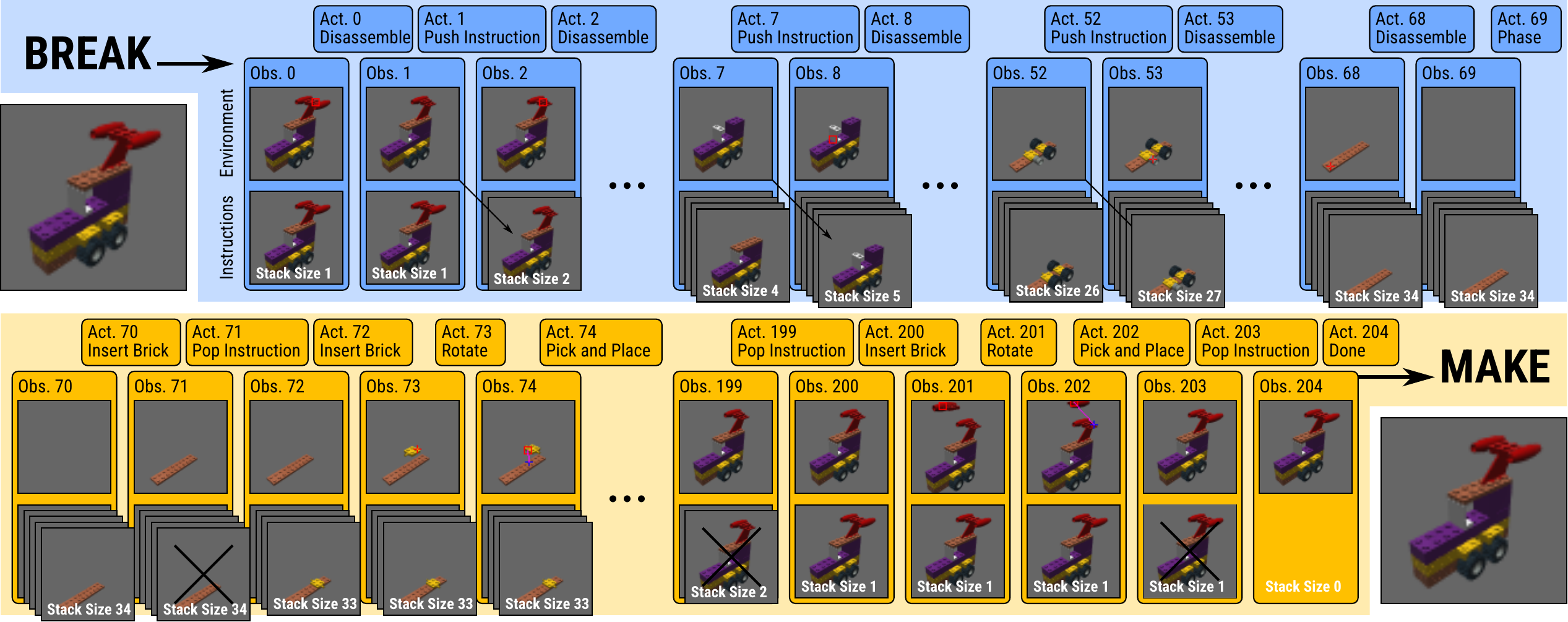}}
  \caption{An example of \ours{} completing the Break and Make task on a previously unseen example from \rcvehicles{}.  Our model saves 34 distinct images to the instruction stack over the first 69 steps.  It then successfully rebuilds the model from scratch using these images over the course of another 135 steps.}
  \label{fig:break_and_make}
  \vspace{-3mm}
\end{figure*}

The Break-and-Make problem is quite challenging, as it requires long-term memory, and interaction with a complex visual environment using a 2D cursor-based action space.  Baseline approaches that use transformers\cite{transformers} and LSTMs\cite{lstms} for long-term memory have struggled to make progress on larger models.  We introduce a new model \textbf{\ours{}} for this problem that uses an explicit memory to store a stack of self-curated instruction images.  \ours{} slowly adds to this memory by iteratively disassembling part of the model, then saving its most recent observation to the top of this instruction stack.  This stack provides the model with a series of short-term visual targets to use when reassembling the model later on, just like a real-world LEGO instruction book.  When rebuilding the model the agent only has to reason about its current observation and the page of the instruction stack that it is currently working on.  Once the agent's current assembly matches this instruction, it can turn the page and get a new short-term target to work towards.  Figure \ref{fig:break_and_make} shows a successful example of this on our new \rcvehicles{} dataset.

In addition to this memory-based model, we also detail several practical components that are either necessary or improve training performance in this space.  These include online imitation learning, conditional action heads and new loss functions for the dense 2D cursor-based action space.

Our primary contributions are:
\begin{enumerate}
    \item We introduce a new instruction-stack based model for the Break and Make problem.
    \item We provide several modeling and training improvements for complex visual action spaces that use a pixel-based cursor.
    \item We provide a new simplified visual interface for the Break and Make problem, along with \rcvehicles{} a new dataset of challenging LEGO models for assembly problems.
    \item Our new model and training recipe achieve much better performance than previous baselines, especially on larger, more complicated models.
\end{enumerate}

%% file: sections/related_work.tex
\section{Related Work}
\subsection{Interactive Understanding and Assembly}
Despite a proliferation of interactive environments with increasing sophistication\cite{robothor, igibson, savva2019habitat, szot2021habitat, multion, yan2018chalet}, long-term structural understanding and assembly problems remain an open challenge in artificial intelligence.  Researchers have explored this in the context of furniture assembly\cite{zhang2023aligning, lim2013parsing}, Minecraft\cite{minedojo}, CAD models\cite{jones2021automate} and robotic assembly problems\cite{zakka2020form2fit}.  In this work, we build on LTRON\cite{ltron}, a recent LEGO simulator designed to provide a building environment for learning agents.  Constructing plans for assembly using a disassembly process has also been explored in the context of multi-part CAD based models\cite{tian2022assemble}, however this work is more concerned with finding collision-free paths through free-space, and does not directly reason about connection points.

LEGO bricks have been a popular substrate for learning assembly problems across a variety of subfields in artificial intelligence.  These include design problems\cite{peysakhov2003using}, robotic assembly\cite{gupta2012duplotrack}, shape reconstruction \cite{kim2014survey, lee2015finding, li2023mobilebrick}, generative modelling\cite{thompson2020building}, and image guided building\cite{chung2021brick, lennon2021image2lego}.  Most similar to our work, Wang et al.\cite{wang2022translating} build LEGO structures from existing instructions.  Our setting is more challenging because the agent must learn to make its own instructions rather than assuming they are already provided.  Furthermore, the action space in LTRON is more difficult as it requires the agent to use a 2D cursor to interact with the scene and contains assemblies with bricks attached to the sides of objects, that cannot be described using simple stacking.

\subsection{Memory}
Memory structures for interactive problems have been studied for decades. Early attempts at implicit (problem-agnostic) memory structures include simple RNNs\cite{rumelhart1985learning, jordan1997serial} and more complex variants such as LSTMs\cite{lstm} and GRUs\cite{gru}.  These methods propagate information forward in time using specialized network architecture.  Neural Turing Machines \cite{graves2014neural} use a learned external memory module to read and write to long term storage.  Recently attention-based transformers\cite{transformers} have become one of the most popular ways to build neural networks that rely on past information to make future decisions.  While transformers are quite effective, they are computationally expensive for large sequences of observations, which limits their use for very long-term memory.

In contrast to implicit memory structures, explicit memory uses some knowledge of the environment or task in order to build a structure more appropriate to the problem.  This can come in the form of geometric or topological map building such as in SLAM\cite{leonard1991simultaneous, choset2001topological, henry2012rgb}, semantic maps in embodied navigation\cite{gordon2019should} or more complex structures that combine multiple components\cite{goat}.  Our approach uses an explicit memory structure designed around the intuitive understanding that assembling a structure can be completed in the reverse order of disassembling it.  This motivates the use of a stack that allows the agent to sequentially build up a series of experiences, and then pop them off one-by-one in reverse order later.

\subsection{Inverse-Graphics} 
Reconstructing geometry and reasoning about 3D structures from images has been an important issue in research fields such as computer-aided design\cite{du2018inversecsg, xu2021inferring} and robotics \cite{chen2023urdformer, suarez2018can, lee2021ikea}.  In particular, prior works such as \cite{willis2021fusion, li2020sketch2cad} use 3D shapes while \cite{zhan2020generative, mo2019structurenet, niu2018im2struct} use images to guide the inverse inference process.  However, when building complex structures such as LEGO models, it is challenging to generate a set of sufficient visual images to predict the reconstruction without dynamically interacting with the object or the environment.

%% file: sections/method.tex
\section{Methods}

\subsection{Environments and Data}
\label{sec:data}
The Break and Make environment in LTRON\cite{ltron} is designed to test an agent's ability to perform complex building tasks on previously unseen target LEGO assemblies.  In this, and many other complex construction problems, no single view of the target object is enough to fully describe it.  Therefore, in order for the model to complete its objective, it must first interactively discover all the components of the LEGO assembly by disassembling it and remembering where all the individual bricks went during that process.  In order to allow for this kind of interactive discovery, the Break and Make problem divides each interactive episode into two distinct phases.  In the first Break phase, the agent is presented with a previously unseen LEGO model and is allowed to interactively inspect and disassemble it, while constructing some long-term memory of the object.  Once it is done, it takes a dedicated Phase Switch action at which point the agent is presented with an empty scene and must use the memory it built in the Break phase to rebuild the model from scratch.

\begin{figure*}[t!]
  \centering
  \makebox[\columnwidth][c]{\includegraphics[width=0.75\textwidth]{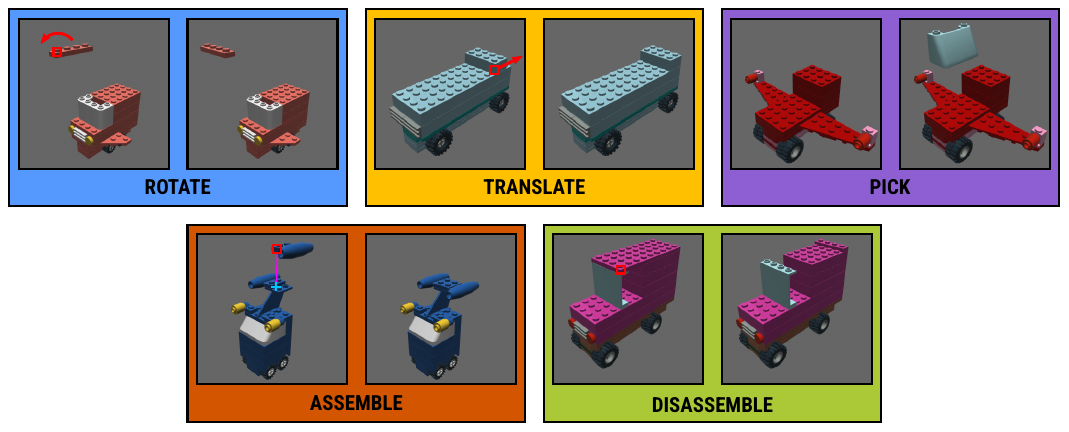}}
  \caption{Our modified LTRON action space without the extra Hand viewport.  We only show the manipulation actions here and do not show the camera rotation and done actions which are unchanged.}
  \label{fig:action_space}
  \vspace{-18pt}
\end{figure*}

In this paper, we use a slightly modified version of the original Break and Make environment with a few changes for simplicity of training.  The original Break and Make environment provided two images as its observation space, one ``table" image representing the current scene, and another ``hand" image representing a single brick that had just been removed from the scene, or was about to be added.  In order to simplify the observation space of our model, we have removed the ``hand" image.  In this updated environment, when a new brick is inserted into the scene, it is simply placed in a floating location above the existing assembly computed using the bounding box of existing bricks, or at the origin if no bricks are present.  We also add a new Translate action mode which allows the agent to shift a brick by a fixed amount by using the cursor to select a connection point and specifying a direction and discrete offset value.  We found this useful for helping the agent recover from small mistakes.

The complete set of action primitives in the new environment are: \textbf{Rotate:} rotates a single brick about a connection point specified by clicking a 2D screen location.  The rotation angle is selected from a set of discrete values in 90 degree increments.  \textbf{Translate:} translates a single brick specified by clicking a 2D screen location.  The translation distance and direction is again selected from a set of discrete values corresponding to multiples of the stud spacing and brick heights.  \textbf{Pick:} inserts a new brick into the scene, by selecting a discrete shape and color index.  The new brick is placed in a floating location above the current assembly.  \textbf{Assemble:} attaches one brick to another by specifying two connection points in the scene.  \textbf{Disassemble:} removes a brick from the scene by specifying a connection point.  All of these actions will only succeed if the operation can be performed without causing a collision.  Figure \ref{fig:action_space} shows the manipulation components of the action space of our updated environment.

Note that despite including collision checking, this environment design does not attempt to simulate real-world physical interaction, as it does not consider forces or other complex dynamics.  This means that the LTRON interface is closer to CAD modelling, where an operator must use a set of discrete tools and a 2D mouse in order to manipulate a virtual 3D object, than a physical manipulation setting that might be encountered in robotics or industrial automation.

We also use a reduced resolution of $128\times128$ pixels for training efficiency.  In addition, we have made a number of improvements to the original LTRON simulator. These updates provide better collision checking and support for a larger number of connection point styles than was available previously. Note that while both the original and our updated environment allow for camera rotation, we found that rotating the camera was not necessary to successfully reassemble the models in the datasets considered here, so we used a fixed camera angle for all experiments.

In order to focus on construction ability and avoid the confounding issues of long-tailed part distributions, we used the 2, 4 and 8 brick random construction assemblies in LTRON and did not train on the Open Model Repository (OMR) data.  Due to the simulator updates mentioned above, we regenerated this random construction data to account for improvements in collision checking. These new assemblies avoid certain rare configurations that resulted in small brick penetrations in the original data.

\begin{figure*}[h]
  \centering
  \makebox[\columnwidth][c]{\includegraphics[width=0.85\textwidth]{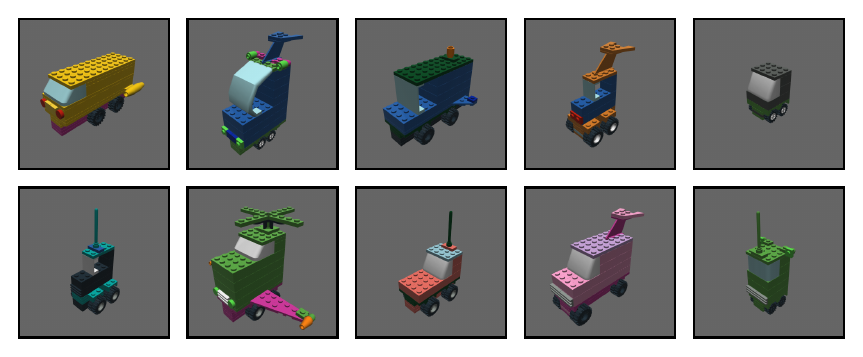}}
  \caption{Examples of the \rcvehicles{} dataset.}
  \label{fig:rcv}
  \vspace{-3mm}
\end{figure*}

In addition to the random construction data, we have also developed a new \rcvehicles{} dataset of randomly constructed vehicles. These were generated with a series of scripted rules defining distributions over the vehicle dimensions, swappable components such as tires and windshield shapes, and optional features such as wings, headlights and helicopter blades. When combined, these distributions have over 21 bits of entropy defining the shape of the vehicle and 29 bits of entropy defining its color combinations. These models vary in size from 19 to 73 bricks, making them substantially larger and more complex than the previous random construction data. Examples of these vehicles can be seen in Figure \ref{fig:rcv}.

Note that both the random brick and the \rcvehicles{} assemblies are not merely top-down stacks of bricks, but require some parts to be placed on the sides, front and back of the constructed object.

\subsection{Instruction Stack}
Our new model \textbf{\ours{}} works by storing an explicit stack of instruction images in order to remember the structure of an assembly at various stages of deconstruction. To do this, we augment the action space discussed in the previous section with two additional \textbf{Push Instruction} and \textbf{Pop Instruction} actions. Push takes the current image from the simulator and adds it to the top of the instruction stack, while Pop removes the top image from the instruction stack.  During training and inference, we do not restrict the size of the instruction stack.

Our learned policy takes in the current image from the simulator as well as the top image of the instruction stack.  During the Break phase when the agent is trying to gather more information about the LEGO assembly, the agent can compare these two images and see if they are similar. If they are, then the agent should disassemble the model further.  Otherwise, the agent should take the Push action to store the new information that has just been gathered.  After completing this process several times, the agent should have an instruction stack with an image of the fully completed assembly on the bottom, and increasingly disassembled images as you move closer to the top.  At the start of the Make phase, the agent will be presented with an empty scene, so the current image from the environment will be empty, while the top of the instruction stack will contain the last brick the model saw during the disassembly process. The agent must then build until the assembly in the current image matches the assembly in the top instruction image.  When they match, the agent can take the Pop action and will then reveal a new instruction image with slightly more of the original assembly remaining.  In this way, the model only needs to reason about two images at a time, which greatly reduces the complexity of the policy.

\subsection{Model}
\label{sec:model}
The learned policy is implemented using a modified vision transformer\cite{vit} with multiple heads that are responsible for different components of the action space.  This model tokenizes the workspace and the instruction images into $16\times16$ pixel patches.  Given that the environment produces images that are $128\times128$ pixels, this results in 64 tokens per image.  The patches from both images are passed through a single linear layer and added to a learned positional encoding.  The model then concatenates a single decoder token and a binary embedding of the current phase (Break or Make) for 130 total tokens.  The transformer consists of 12 blocks with 512 channels and 8 heads each.
 
\begin{figure}[t!]
  \centering
  \makebox[\columnwidth][c]{\includegraphics[width=0.65\columnwidth]{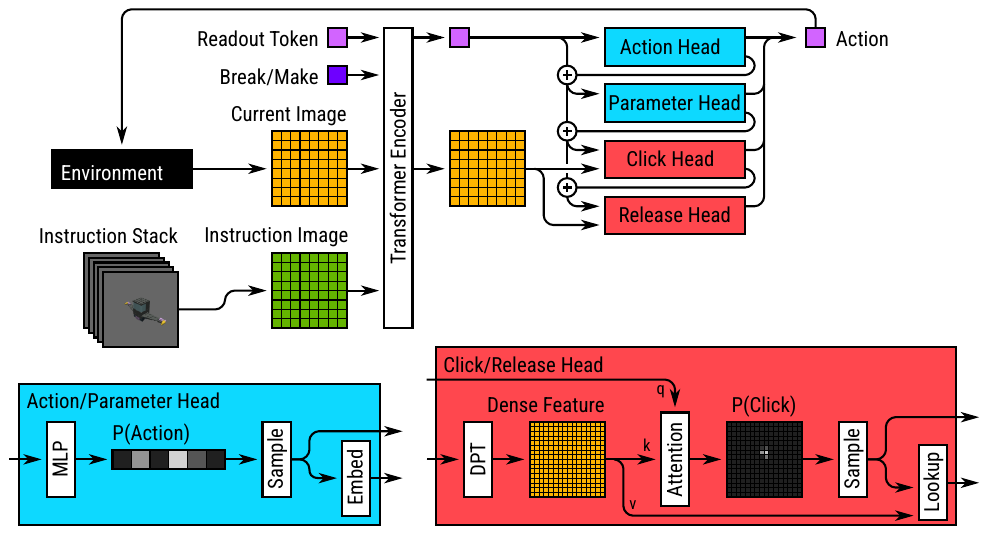}}
  \caption{Architecture of the \ours{} model.  The current image from the environment, and the top image of the instruction stack are tokenized and provided as input to a vision transformer encoder, along with a single readout token and another discrete token that indicates whether the current phase is Break or Make.  The readout token's feature decodes a series of discrete action and parameter heads that determine the high level action mode (Rotate/Translate/Pick/Assemble/Disassemble) as well as action parameters such as the rotation angle or translate distance and direction.
  The cursor click and release locations are sampled from an attention map
  comparing
  features from a DPT decoder.
  \vspace{-18pt}
  }
  \label{fig:architecture}
\end{figure}

To compute an action, the output of the decoder token is then passed through a set of decoder heads to predict distributions for the action mode such as Disassemble, Assemble, or Rotate and mode-specific parameters such as Rotate Direction when performing a Rotate action or Brick Shape and Color when inserting a new brick.
In order to predict 2D cursor click and release locations, the 64 output tokens of transformer blocks 3,6,9 and 12 that correspond to the current image are combined using two separate DPT\cite{dpt} decoders to produce dense feature maps at the resolution of the original input image.  We found it beneficial to condition these click locations on the high level action and parameters sampled from the initial decoder heads.  This is accomplished by passing the sampled high level actions to an embedding layer and adding the resulting feature to the output of the decoder token.  This value is then used to compute a distribution over click locations using dot product attention over the dense features computed by the DPT decoder.  This conditional structure is similar to models used in game AI with complex action spaces \cite{vinyals2019grandmaster}.  Figure \ref{fig:architecture} shows our policy model and how these components fit together.

\subsection{Online Training}
We trained our model using online imitation learning similar to DAgger\cite{dagger} and show ablations using behavior cloning as well.  To do this we built a fast expert that can provide online supervision for trajectories generated by the learning model during training.  This is in contrast to \cite{ltron} which used an expert that was too slow for online labeling. 
Note that while our new expert is much faster, it is not able to construct plans in cases where the agent makes too many mistakes or deviates too far from the target assembly.  In these cases, we simply terminate the training episode early.  When there are multiple possible best actions that the expert could suggest, one of them is selected at random.  To avoid ambiguity, our expert instructs the agent to push an image to the instruction stack each time a brick is removed during the break phase.

The online training algorithm alternates between generating new data by acting in the environment according to either the expert or the learning model, and then training on a randomized subset of the data generated over the past several iterations.  When generating data, a fixed percentage of the environment steps are generated by sampling actions according to the expert, and the rest are sampled by acting according to the agent.  We refer to this expert mixture constant as $\alpha$, and for our main experiments, we found that $\alpha=0.75$, representing a mix of $75\%$ expert-generated and $25\%$ model-generated data worked well.  Note that regardless of which model is controlling the simulator, the expert's actions are always used for supervision.  Incorporating trajectories generated by the agent in this way allows the model to learn to recover from its mistakes: when the model takes an inappropriate action it will reach part of the state space that would not have been encountered if acting according to the expert, yet seeing the expert's advice in these states shows the model how to correctly recover from this behavior.

As noted by Czarnecki et al. \cite{czarnecki2019distilling}, this method of using the model to generate data with direct supervision from the expert can be unstable.  This instability occurs frequently when the data distribution shifts as the model gets better at some parts of the state space and learns to take different actions.  This problem is exacerbated in settings with long trajectories where many nearby frames appear similar to each other.  If proper care is not taken, the model can quickly overfit to the data it has most recently seen and catastrophically forget prior examples.  The original DAgger algorithm addresses this by periodically retraining the entire model on all data collected so far.  Unfortunately, this is somewhat impractical for long training runs with millions of steps.  To mitigate this issue we maintain a replay buffer and train on data randomly sampled from this experience.  See the supplementary material for the pseudo-code we use to train our model.

\subsection{Cursor Losses}
\label{sec:cursor_losses}
We experimented with several approaches to computing losses for the cursor click and release locations.  During online training, we would like the click-and-release decoders to each produce a distribution over locations that can be sampled in order to generate a variety of training data.  To compute the probability of clicking on a particular pixel $p(i,j)$, the dense raw value $x_{i,j}$ predicted at that pixel location is normalized using a standard softmax:
\[
p(i,j) = \frac{\exp(x_{i,j}) }{ \sum_{i',j'} \exp(x_{i',j'})}
\]
When the expert's action suggests clicking on a particular LEGO connection point in the scene, there are usually multiple ``acceptable" pixels that correspond to the same connection point which complicates the choice of loss function.

One option is to use binary cross-entropy loss using the mask of acceptable pixel locations as a target.  This assumes a different probabilistic interpretation of the output, one where multiple pixels can be chosen at the same time instead of just one, but still may be a useful way to encourage the model to put a high probability on the acceptable pixels.  This loss function encourages the model to increase the probability of all acceptable pixels, without considering the cross-pixel relationships.

Another option commonly used in keypoint detection is to construct target heatmaps using a small Gaussian distribution around correct locations and supervise the output values using a mean-squared-error loss \cite{bulat2016human}.  We opted against this as it does not lend itself well to softmax sampling, and may add probability mass outside the desired pixel boundaries.  However, we can use the mean-squared-error loss to simply push all acceptable pixels toward a large positive constant, and all unacceptable pixels towards a large negative constant.  Here we used a constant such that if only one pixel in the image assumed the positive constant, and all others assumed the negative constant, the probability of selecting the single pixel in the softmax would be 0.999.  For $128\times128$ pixel images, these constants are $\pm 8.3$.

Finally, we consider a loss function in which the probability of acceptable pixels is summed and the probability of unacceptable pixels is summed forming a new two-way distribution.  We then supervise this new distribution to maximize the probability of choosing an acceptable pixel using cross-entropy.
\[
L = -\log \sum_{i,j} y_{i,j} \frac{\exp x_{i,j}}{\sum_{i',j'} \exp x_{i',j'}}
\]
This allows the model to place probability mass on any of the acceptable pixels while decreasing the probability mass of all unacceptable pixels.  We find that this loss function outperforms the others and discuss this in Section \ref{sec:experiments_loss}.

%% file: sections/experiments.tex
\section{Experiments}

\subsection{Evaluation}
\label{sec:evaluation}
We use the metrics from \cite{ltron} to evaluate the quality of our learned agents.  The first ($F1_b$) is an $F1$ score over the brick shape and color, ignoring pose.  The second ($F1_a$) is over the brick shape, color, and pose after computing the best single rigid transformation to bring the estimated and ground-truth assemblies into alignment with each other.  Assembly Edit Distance ($AED$) measures how many rigid transforms (edits) are required to bring all the bricks in the estimated assembly into alignment with the bricks in the ground-truth assembly, with additional penalties for extra and missing bricks.  The final ($F1_e$) is an $F1$ score over the edges (connections) between bricks after using the alignment computed in $AED$ to construct a mapping between bricks in the estimated assembly and bricks in the ground-truth assembly.

\subsection{Break and Make}
\label{sec:break_and_make}
To evaluate the effectiveness of the \ours{} model, we trained it on our modified version of the Break and Make task using the randomly constructed assemblies and vehicles discussed in Section \ref{sec:data}.  Table \ref{tab:break_and_make} shows the performance of our model on these datasets compared to the reported numbers of the LSTM, Studnet-A, and Studnet-B baselines from \cite{ltron}.  The Studnet models are causal transformers that take in a series of deduplicated tiles from the beginning of an episode until the current time step and use a series of decoders to make predictions for the current action.  All three of these models require a long horizon to make decisions, while \ours{} requires only the current frame and the top of the instruction stack.  See \cite{ltron} for details of the Studnet and LSTM models.  Note that due to the updated observation space and the small changes to the random construction data due to the improved collision checker, these models should not be considered to have been trained in the same environment, and so the comparisons are only approximate.  Regardless of these differences, the \ours{} model is able to reconstruct large models with much greater accuracy than was previously possible, and the large performance gap clearly demonstrates a new level of capability.  Note that we were not able to train the LSTM and Studnet methods from \cite{ltron} on the new \rcvehicles{} data as they require the entire history of past frames to make each decision.  The \rcvehicles{} assemblies can take over 150 steps to correctly disassemble and reassemble, which are much larger sequences than we could effectively train on available hardware.

\begin{table}[h]
\vspace{-5pt}
\centering
\begin{tabularx}{0.75\columnwidth}{Xcccc}
 \toprule
 RC-2* & {\small $F1_b\uparrow$} & {\small $F1_e\uparrow$} & {\small $F1_a\uparrow$} & {\small $AED\downarrow$} \\
 \midrule
 \textbf{\ours{}} & \textbf{0.98} & \textbf{0.95} & \textbf{0.93} & \textbf{0.18} \\
 LSTM & 0.61 & 0.38 & 0.43 & 2.16 \\
 Studnet-A & 0.90 & 0.86 & 0.58 & 1.11 \\
 Studnet-B & 0.87 & 0.77 & 0.57 & 1.30 \\
 \midrule
 RC-4* & {\small $F1_b\uparrow$} & {\small $F1_e\uparrow$} & {\small $F1_a\uparrow$} & {\small $AED\downarrow$} \\
 \midrule
 \textbf{InstructioNet} & \textbf{0.80} & \textbf{0.69} & \textbf{0.71} & \textbf{2.39} \\
 LSTM & 0.41 & 0.09 & 0.13 & 7.25 \\
 Studnet-A & 0.56 & 0.29 & 0.24 & 5.80 \\
 Studnet-B & 0.64 & 0.34 & 0.25 & 5.48 \\
 \midrule
 RC-8* & {\small $F1_b\uparrow$} & {\small $F1_e\uparrow$} & {\small $F1_a\uparrow$} & {\small $AED\downarrow$} \\
 \midrule
 \textbf{InstructioNet} & \textbf{0.68} & \textbf{0.62} & \textbf{0.63} & \textbf{6.30} \\
 LSTM & 0.02 & 0.00 & 0.02 & 16.05 \\
 Studnet-A & 0.02 & 0.01 & 0.01 & 15.87 \\
 Studnet-B & 0.38 & 0.14 & 0.12 & 13.90 \\
 \midrule
 \rcvehicles{} & {\small $F1_b\uparrow$} & {\small $F1_e\uparrow$} & {\small $F1_a\uparrow$} & {\small $AED\downarrow$} \\
 \midrule
 \textbf{InstructioNet} & \textbf{0.59} & \textbf{0.51} & \textbf{0.53} & \textbf{43.36} \\
 \bottomrule
\end{tabularx}
\vspace{5pt}
\caption{\ours{} compared against the LSTM and Studnet baselines from \cite{ltron}.  See Section \ref{sec:evaluation} for details on these metrics, and \ref{sec:break_and_make} for an important note on the direct comparability of these methods.}
\label{tab:break_and_make}
\vspace{-40pt}
\end{table}

\subsection{Online Training}
\label{sec:online_training}
In order to show the effectiveness of online training using sequences of actions and observations generated by the learning model, we also train a model on sequences generated only by the expert on the RC-2 and RC-4 datasets.  To do this, we set the expert mixture ($\alpha$) to $1.0$, which is equivalent to behavior cloning.
The results are shown in the \textbf{Online Training} section of Table \ref{tab:ablations}.
While these models underperform relative to the default expert mixture ($\alpha = 0.75$) that includes training from an online expert, it still shows that significant progress can be made with offline training on this problem.

\begin{table}[t]
\centering
\begin{tabularx}{0.75\columnwidth}{Xcccc}
 \toprule
 \multicolumn{5}{l}{\textbf{Online Training} (Section \ref{sec:online_training})} \\
 \midrule
 RC-2 & {\small $F1_b\uparrow$} & {\small $F1_e\uparrow$} & {\small $F1_a\uparrow$} & {\small $AED\downarrow$} \\
 \midrule
 $\bm{\alpha=0.75}$ & \textbf{0.98} & \textbf{0.95} & \textbf{0.93} & \textbf{0.18} \\
 $\alpha=1.0$ & 0.97 & 0.93 & 0.90 & 0.29 \\
 \midrule
 RC-4 & {\small $F1_b\uparrow$} & {\small $F1_e\uparrow$} & {\small $F1_a\uparrow$} & {\small $AED\downarrow$} \\
 \midrule
 $\bm{\alpha=0.75}$ & \textbf{0.80} & \textbf{0.69} & \textbf{0.71} & \textbf{2.39} \\
 $\alpha=1.0$ & 0.77 & 0.68 & 0.66 & 2.88 \\
 \midrule
 \multicolumn{5}{l}{\textbf{Loss Functions} (Section \ref{sec:experiments_loss})} \\
 \midrule
 RC-2 & {\small $F1_b\uparrow$} & {\small $F1_e\uparrow$} & {\small $F1_a\uparrow$} & {\small $AED\downarrow$} \\
 \midrule
 Summed CE & \textbf{0.98} & \textbf{0.95} & \textbf{0.93} & \textbf{0.18} \\
 BCE & 0.91 & 0.72 & 0.72 & 0.92 \\
 MSE & 0.91 & 0.50 & 0.66 & 1.00 \\
 \midrule
 \multicolumn{5}{l}{\textbf{Conditional Actions} (Section \ref{sec:conditional_action_generation})} \\
 \midrule
 RC-2 & {\small $F1_b\uparrow$} & {\small $F1_e\uparrow$} & {\small $F1_a\uparrow$} & {\small $AED\downarrow$} \\
 \midrule
 1.7M & \textbf{0.98} & 0.94 & 0.92 & 0.22 \\
 \textbf{2.6M} & \textbf{0.98} & \textbf{0.95} & \textbf{0.93} & \textbf{0.18} \\
 Cut 1.7M & 0.97 & 0.89 & 0.84 & 0.43 \\
 Cut 2.6M & \textbf{0.98} & 0.00 & 0.50 & 1.09 \\
 \midrule
 \multicolumn{5}{l}{\textbf{Selective Modification} (Section \ref{sec:selective_modification})} \\
 \midrule
 RC-2 & {\small $F1_b\uparrow$} & {\small $F1_e\uparrow$} & {\small $F1_a\uparrow$} & {\small $AED\downarrow$} \\
 \midrule
 Original & 0.98 & 0.95 & 0.93 & 0.18 \\
 Altered Color & 0.98 & 0.93 & 0.95 & 0.21 \\
 \midrule
 RC-4 & {\small $F1_b\uparrow$} & {\small $F1_e\uparrow$} & {\small $F1_a\uparrow$} & {\small $AED\downarrow$}\\
 \midrule
 Original & 0.80 & 0.69 & 0.71 & 2.39 \\
 Altered Color & 0.78 & 0.67 & 0.68 & 2.63 \\
 \bottomrule
\end{tabularx}
\vspace{5pt}
\caption{The results of various ablations, see the sections listed above for details.}
\label{tab:ablations}
\vspace{-24pt}
\end{table}

\subsection{Loss Functions}
\label{sec:experiments_loss}
We evaluate the effectiveness of our cursor loss function by comparing it against the binary cross entropy and constant regression methods discussed in Section \ref{sec:cursor_losses}.  We find that even on the relatively easy two-brick models, the summed-probability loss outperforms these other techniques.
The results are shown in the \textbf{Loss Functions} section of Table \ref{tab:ablations}.
Note that due to the difference in magnitude between these losses, we adjusted the learning rate for these methods in an attempt to achieve the best results possible.  We found that both benefited from a higher learning rate of $5\times10^{-4}$ rather than the default $5\times10^{-5}$ used for the summed-probability loss.

\subsection{Conditional Action Generation}
\label{sec:conditional_action_generation}
We also test the importance of sequentially conditioning the action heads on one another as discussed in Section \ref{sec:model} by training a new model where these conditional connections are cut.  This corresponds to cutting the magenta connections coming out of the Action Head, Parameter Head and Click Head in Figure \ref{fig:architecture}.  We found that cutting these connections leads to training instability where after a certain point the model loses its ability to effectively use the cursor to connect bricks together.  In light of this, we report results after 1.7M frames, right before the instability occurs in addition to the default 2.6M frames after the instability occurs.  We also report an evaluation of the default model at 1.7M frames for comparison.  Note that even before training became unstable, the model without the conditional connections was significantly underperforming the default model.
The \textbf{Conditional Actions} section of Table \ref{tab:ablations} shows these results.

\subsection{Selective Modification}
\label{sec:selective_modification}
We also tested our model on a new task that requires the agent to rebuild the model with one of the brick colors altered.  In this setting, the model receives two additional tokens, one which specifies the color to change, and the other that specifies a new color.  The agent's objective is to rebuild the model with all bricks using the original color instead built with the new color.
The \textbf{Selective Modification} section of Table \ref{tab:ablations} shows the performance on this task for two and four brick models.
While, performance degrades slightly on this problem, the model still performs quite well.  This demonstrates the model's ability to reason not only about reproducing the exact same model seen during the break phase, but also incorporating new instructions when rebuilding.

\subsection{Hyperparameters}
Unless otherwise mentioned, we used AdamW with a learning rate of $5\times10^{-5}$, $\beta_1=0.9$, $\beta_2=0.95$ and weight decay of 0.1.  For RC-8 and RC-V, the learning rate was cut to $1\times10^{-5}$ after 15.8M frames.  All models were trained on a single Nvidia 4090 or A40 graphics card.  The larger RC-8 and \rcvehicles{} runs were trained on 19.7M training steps which took five consecutive days per run.  The RC-2 and RC-4 datasets were trained on 2.6M and 7.9M steps respectively which took between one and three consecutive days per run.  Table \ref{tab:hyperparameters} shows the training hyperparameters used to train the models for each dataset.

\begin{table}[h]
\vspace{-8pt}
\centering
\begin{tabularx}{0.75\columnwidth}{Xcccc}
\toprule
 Hyperparameters & RC-2 & RC-4 & RC-8 & RC-V \\
 \midrule
 Total Training Steps & 2.6M & 7.9M & 19.7M & 19.7M \\
 New Data Steps Per Epoch & 8K & 8K & 8K & 32K \\
 Training Steps Per Epoch & 16K & 16K & 16K & 65K \\
 Replay Buffer Size & 32K & 32K & 32K  & 131K \\
 Expert Data Mixture ($\alpha$) & 0.75 & 0.75 & 0.75 & 0.75 \\
 \bottomrule
\end{tabularx}
\vspace{5pt}
\caption{Hyperparameters used to train the different datasets.}
\label{tab:hyperparameters}
\vspace{-30pt}
\end{table}

\subsection{Qualitative Evaluation}
\label{sec:qualatative}
\begin{figure*}[h]
  \vspace{-8pt}
  \centering
  \makebox[\columnwidth][c]{\includegraphics[width=0.85\textwidth]{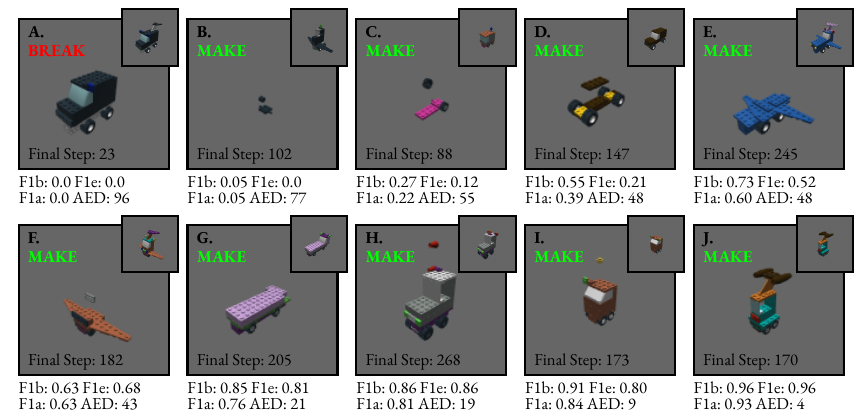}}
  \caption{Examples of \ours{} reconstructions trained on the \rcvehicles{} dataset.  The top right overlay shows the target assembly.  These examples were chosen to present a diverse array of failure and success cases.  See Section \ref{sec:qualatative} for descriptions of these failures and Section \ref{sec:evaluation} for an explanation of the evaluation metrics.}
  \label{fig:reconstruction_examples}
  \vspace{-32pt}
\end{figure*}

Figure \ref{fig:reconstruction_examples} shows ten representative failure and success cases of our model on the \rcvehicles{} dataset sorted by their F1a score.  Example A shows a case where the model fails to complete the Break phase due to a small ornament on top of the car that the model does not realize it needs to remove.  In examples B, C, and D the model successfully completes the Break phase, but then struggles to complete the early part of the model.  In example E, the model initially placed the wings correctly, but then misclicked as it was placing a later piece and inadvertently moved one of the wings to the wrong location and it was not able to recover.  In example F, the model incorrectly built the front grille and fails to either undo its mistakes or move on.  In example G, the model misplaced one brick in the back of the car and was also not able to correct its mistake.  Examples H and I show cases where the model almost completely reconstructs the assembly, but gets hung up on the small ornamental details on the roof.  Finally, example J shows an almost perfect reconstruction where one hidden brick is incorrect.

%% file: sections/conclusion.tex
\section{Conclusion}
We have demonstrated substantially improved performance over previous baselines on the Break and Make problem, using a model with explicit instruction memory.  The failure modes of this approach suggest that performance could be improved by working on solutions that avoid getting stuck, and that find ways to push forward even if it means making a local mistake.  While our approach is successful, it requires an online expert which can provide explicit instructions not only for the inspection and reconstruction process but also for the process of storing and retrieving memory.  This limits the utility of this method in real-world settings, where an online expert may not be available.  The \ours{} approach may not be appropriate for problems that do not follow our assumption that assembly can be completed by approximately reversing the disassembly process.  Nevertheless the advantage of this method over prior approaches which considered the entire observation history points to the effectiveness of considering only a portion of memory at a time when making decisions.

%% file: appendices/additional_ablations.tex
\section{Additional Ablations}

\subsection{Camera Motion}
We also test the extent to which direct image comparison enables our results.  To evaluate this, we retrained our model on the RC-V dataset, but with the camera in each frame rotated about the center of the assembly by $\pm 0.1$ radians, and translated by $\pm 10$ LDU in X, Y and Z.  This means that when comparing the current simulator image with the top image of the construction stack, they will be viewed from slightly different viewpoints, and will not be aligned pixel-by-pixel.  For reference, we also report numbers for the original model that was trained without camera motion evaluated on data with camera motion.  The results of this experiment is shown in Table \ref{tab:camera_motion}.  Clearly these shifts in viewpoint negatively impact performance in a substantial way even when retrained on this data (see row Motion(T)), though the model is still able to achieve some success.  We hypothesize that this decrease in performance is due to the difficulty of accurately comparing small offsets at low resolution.  The problem becomes much worse when evaluating the model that was not trained with camera motion on data that contains camera motion (see row Motion(Unt)).  In this case the model completely fails given that it was only ever trained on neatly aligned images.
\begin{table}[h]
\centering
\begin{tabularx}{0.75\columnwidth}{Xcccc}
 \toprule
 RC-V & {\small $F1_b\uparrow$} & {\small $F1_e\uparrow$} & {\small $F1_a\uparrow$} & {\small $AED\downarrow$} \\
 \midrule
 \textbf{No Camera Motion} & \textbf{0.59} & \textbf{0.51} & \textbf{0.53} & \textbf{46.36} \\
 Camera Motion (Trained) & 0.45 & 0.29 & 0.33 & 56.37 \\
 Camera Motion (Untrained) & 0.00 & 0.00 & 0.00 & 78.75 \\
 \bottomrule
\end{tabularx}
\vspace{5pt}
\caption{The default training approach (No Motion) compared against a model trained under small camera motion (Motion(Trained)) and the model trained on no motion, but evaluated under camera motion (Motion (Untrained)).}
\vspace{-20pt}
\label{tab:camera_motion}
\end{table}

\subsection{Expert Instruction Images}
Given that our model only requires the current simulator image, the top image of the instruction stack and the current phase, we also explore the success of the model in a setting where we use the expert to generate the instruction stack during the break phase, but then do assembly with a learned agent using the expert's instruction stack as input.  Note that the models used here are not retrained, but use the same checkpoints from our main results.  The improvement in performance when using the expert instructions aligns with the fact that the model sometimes fails to complete the Break phase on its own.  Having access to the expert instructions and starting each episode in the Make phase removes these failure cases.  Unsurprising the gap shrinks for smaller models where the overall performance is higher.  While it is theoretically possible that the distribution of expert instructions differs from that which is typically produced by the agent, to the extent that this distribution shift exists, it does not appear to be hindering the results here.

\begin{table}[h]
\centering
\begin{tabularx}{0.75\columnwidth}{Xcccc}
 \toprule
 RC-2 & {\small $F1_b\uparrow$} & {\small $F1_e\uparrow$} & {\small $F1_a\uparrow$} & {\small $AED\downarrow$} \\
 \midrule
 Model generated instructions & 0.98 & 0.95 & 0.93 & 0.18 \\
 \textbf{Expert generated instructions} & \textbf{0.99} & \textbf{0.96} & \textbf{0.96} & \textbf{0.17} \\
 \midrule
 RC-4 & {\small $F1_b\uparrow$} & {\small $F1_e\uparrow$} & {\small $F1_a\uparrow$} & {\small $AED\downarrow$} \\
 \midrule
 Model generated instructions & 0.80 & 0.69 & 0.71 & 2.39 \\
 \textbf{Expert  generated instructions} & \textbf{0.86} & \textbf{0.77} & \textbf{0.76} & \textbf{1.94} \\
 \midrule
 RC-8 & {\small $F1_b\uparrow$} & {\small $F1_e\uparrow$} & {\small $F1_a\uparrow$} & {\small $AED\downarrow$} \\
 \midrule
 Model generated instructions & 0.68 & 0.62 & 0.63 & 6.30 \\
 \textbf{Expert generated instructions} & \textbf{0.83} & \textbf{0.75} & \textbf{0.75} & \textbf{4.39} \\
 \midrule
 RC-V & {\small $F1_b\uparrow$} & {\small $F1_e\uparrow$} & {\small $F1_a\uparrow$} & {\small $AED\downarrow$} \\
 \midrule
 Model generated instructions & 0.59 & 0.51 & 0.53 & 46.36 \\
 \textbf{Expert generated instructions} & \textbf{0.71} & \textbf{0.63} & \textbf{0.65} & \textbf{33.81} \\
 \bottomrule
\end{tabularx}
\vspace{5pt}
\caption{The fully trained model evaluated on instructions generated by the model and by the expert.}
\vspace{-20pt}
\label{tab:expert_instructions}
\end{table}

%% file: appendices/expert_details.tex
\section{Online Expert Details}
Here we include details on the logical procedure used to generate expert actions.  At each time step, the expert has access to the current assembly $\hat{A}$ and the target assembly $A$.  We also keep track of a list of assemblies $\bar{A}$ that correspond to each time an instruction image was pushed onto the instruction stack.  Algorithm \ref{algo:expert} shows the procedure for generating expert actions.  We use the shorthand $|A|$ to refer to the number of bricks in an assembly.  The subroutine $\text{matching}(A,B)$ computes the single transform that best aligns the assemblies $A$ and $B$ and creates a lookup table between bricks that match under the alignment.  The subroutine $\text{matching\_statistics}(m,A,B)$ takes a matching and a lookup table $m$ and two assemblies and computes a set of true positives $t_p$, false positives $f_p$ and false negatives $f_n$.  It also returns the disconnected true positives $d_p$ which are bricks in the estimated assembly that match the shape and color of a brick in the ground truth assembly, but is not in the correct pose and is not connected properly, as well as connected true positives $c_p$ which are bricks that match the shape and color of a brick in the ground truth assembly, but are not in the correct pose, but do have at least one connection point connected correctly.  The \textbf{REMOVE}, \textbf{ROTATE}, and \textbf{ASSEMBLE} require that an appropriate connection point is visible for cursor selection.  If one is not, \textbf{TERMINATE EARLY} is returned instead.

\begin{algorithm}
\caption{Online Expert}
\label{algo:expert}
\begin{algorithmic}
\Require Current Assembly $\hat{A}$
\Require Target Assembly $A$
\Require Stack of Instruction Assemblies $\bar{A}$
\Require Current Phase $p$
\State{Compute assembly matching $m,T = \text{match}(\bar{A}_{top},\hat{A})$}
\State{$t_p,d_p,c_p,f_p,f_n = \text{match\_statistics}(m,\bar{A}_{top}, \hat{A})$}

\State $m_p = d_p \cup c_p$
\If {$|f_n| > 1$ \textbf{or} $|m_p| > 1$ \textbf{or} ($|m_p|$ \textbf{and} $|A| \neq |\bar{A}_{top}|$)}
    \State Return \textbf{TERMINATE EARLY}
\EndIf

\If{$p = \text{Break}$}
    \State $r = |\bar{A}_{top}| - |\hat{A}|$
    \If{$r > 1$ \textbf{or} $r < 0$}
        \State Return \textbf{TERMINATE EARLY}
    \ElsIf{$r = 1$}
        \State Return \textbf{PUSH INSTRUCTION}
    \ElsIf{$|\hat{A}| = 0$}
        \State Return \textbf{SWITCH TO MAKE PHASE}
    \EndIf
\Else
    \If{$\hat{A} = \bar{A}_{top}$}
        \State Return \textbf{POP INSTRUCTION}
    \ElsIf{$\hat{A} = A$}
        \State Return \textbf{DONE}
    \EndIf
\EndIf

\If{$|f_p| > 0$}
    \State Return \textbf{REMOVE($f_p$)}
\ElsIf{$|f_n| > 0$}
    \State Return \textbf{INSERT($f_n$)}
\ElsIf{$|c_p| > 1$}
    \State Return \textbf{ROTATE($c_p$)}
\ElsIf{$|d_p| > 1$}
    \If{$\text{orientation\_correct}(d_p)$}
        \State Return \textbf{ASSEMBLE($d_p$)}
    \Else
        \State Return \textbf{ROTATE($d_p$)}
    \EndIf
\EndIf

\end{algorithmic}
\end{algorithm}

During the Break phase, this expert provides advice that removes one brick at a time, then pushes a new instruction image until no more bricks remain.  It then switches to the make phase.  During the Make phase, the expert will add a new brick and move it into place until the current assembly matches the assembly that was stored with the top instruction image, then pop that image off the stack.  When rolling out using the learning agent, the expert is capable of recovering from incorrectly inserted bricks (by removing them), incorrectly placed bricks (by moving them).  However it is not able to handle situations where the current assembly differs from the assembly corresponding to the top of the instruction stack by two or more bricks.

%% file: appendices/training_algorithm.tex
\section{Online Training Algorithm}
Our online training algorithm mixes off-policy data generated by the expert with on-policy data generated by the learner.  The percentage of expert-generated data is controlled by a constant $\alpha$ which is set to 0.75 by default ($75\%$ expert generated data).  In both scenarios, the expert's advice is used a label regardless of which policy was used to construct actions in the environment.

The algorithm maintains a fixed-capacity replay buffer $D$ of recent experience.  Training proceeds in epochs, in which a fixed number of new dataset transitions $N_{steps/epoch}$ are generated according to the expert ratio $\alpha$.  These transitions are added to the replay buffer, evicting the oldest data.  The model is then trained for a fixed number of steps $N_{train/epoch}$ on data sampled uniformly from the replay buffer.  This process repeats until a maximum number of steps $N_steps$ have taken in the environment.  In practice, online data generation was parallelized across 32 copies of the environment.  See Algorithm \ref{algo:online_training} for step-by-step psuedocode.

\begin{algorithm}
\caption{Online Training}
\label{algo:online_training}
\begin{algorithmic}
\Require LTRON Environment $E$
\Require Expert $\pi^*$
\Require Total Steps $N_{steps}$
\Require Epoch Rollout Steps $N_{steps/epoch}$
\Require Epoch Train Steps $N_{train/epoch}$
\Require Max Dataset Size $C$
\Require Expert-guided percentage $\alpha$
\State Initialize policy $\pi_\theta$.
\State Initialize $D = \left\{ \right\}$.
\While {$i < N_{steps}$}
    \For {$j=1$ \textbf{to} $N_{steps/epoch}$}
        \State Sample action $a^* \sim \pi^*(o)$
        \State Sample action $a \sim \pi_\theta(o)$
        \If {$j/N_{steps/epoch} < \alpha$}
            \State Execute action $o = E(a^*)$
        \Else
            \State Execute action $o = E(a)$
        \EndIf
        \State Add $(o,a^*)$ to $D$
        \State If $|D| > C$, evict the oldest entry
    \EndFor
    \For {$j=1$ \textbf{to} $N_{train/epoch}$}
        \State Sample a batch $(o,a^*) \sim D$
        \State Train $\pi_\theta(o) \rightarrow a$
    \EndFor
    \State $i=i+N_{steps/epoch}$
\EndWhile
\end{algorithmic}
\end{algorithm}